\documentclass[10pt,twocolumn]{article}

\usepackage{times}
\usepackage{epsfig}
\usepackage{graphicx}
\usepackage{amsmath}
\usepackage{amssymb}

\usepackage[pagebackref=true,breaklinks=true,letterpaper=true,colorlinks,bookmarks=false]{hyperref}

\begin{document}

\title{High-Order Nonparametric Belief-Propagation for Fast Image Inpainting}

\author{Julian~John~McAuley, Tib\'erio~S.~Caetano%
\thanks{J.~McAuley and T.~Caetano are with NICTA's Statistical Machine Learning program, Locked Bag 8001 Canberra ACT 2601 Australia, and the Research School of Information Sciences and Engineering, Australian National University, ACT 0200, Australia. e-mail: \texttt{Julian.McAuley@rsise.anu.edu.au}, \texttt{Tiberio.Caetano@nicta.com.au}}}%

\maketitle

\begin{abstract}
In this paper, we use belief-propagation techniques to develop fast algorithms for image inpainting. Unlike traditional gradient-based approaches, which may require many iterations to converge, our techniques achieve competitive results after only a few iterations. On the other hand, while belief-propagation techniques are often unable to deal with high-order models due to the explosion in the size of messages, we avoid this problem by approximating our high-order prior model using a Gaussian mixture. By using such an approximation, we are able to inpaint images quickly while at the same time retaining good visual results.
\end{abstract}

\section{Introduction}
\label{sec:introduction}

In order to restore a corrupted image, one needs a model of how uncorrupted (i.e.\ natural) images appear. In the Markov random field Bayesian paradigm for image restoration \cite{Geman84stochastic}, natural images are modeled via an \emph{image prior}. This is a probabilistic model that encodes how natural images behave \emph{locally} (in the vicinity of every pixel). An inference algorithm is then used to restore the image, whose aim is to find a consensus among all vicinities on which global solution is most compatible with a natural image (while still similar to the corrupted image---a corrupted tree should be restored as an uncorrupted tree, not as an uncorrupted house). The simplest example of an image prior is perhaps the pairwise model presented in \cite{Geman84stochastic} -- which simply expresses that neighboring pixels are likely to share similar gray-levels. However, it is easy to see that such a model fails to capture a great deal of important information about natural images. For example, `edges' are highly penalized by such a prior, and it is unable to encode any information about `texture'. Only by using higher-order priors will one be able to capture this important information.

An example of a high-order prior is the \emph{field of experts} model \cite{roth05fields}, which is parameterized as the product of a selection of filters (or `experts'). Each of these filters is typically a patch of $3\times 3$ or $5\times 5$ pixels, resulting in a $9$ or $25$-dimensional prior respectively. Unfortunately, such a high-dimensional prior limits the practicality of many inference algorithms. Even though it may be possible to use smaller (for example, $2\times 2$) patches \cite{lanroth}, we are still limited by the number of gray-levels used to properly represent natural images (typically $256$). Updating a single pixel using a Gibbs sampler (for example) requires us to consider all $256$ possible gray-levels. Since Gibbs samplers may typically take hundreds (or thousands) of iterations to converge, they are simply impractical in this setting.

While belief-propagation techniques tend to converge in fewer iterations \cite{yedidia00generalized}, they are often equally impractical. Since adjacent cliques may share as many as $6$ nodes (using a $3\times 3$ model), the size of the messages passed between them may be as large as $256^6$. Even if we only use $2\times 2$ cliques, the size of our messages may still be as large as $256^2$, which remains impractical for many purposes. Although inpainting has been previously approached using belief-propagation techniques in \cite{levin03}, they do not deal with $2\times 2$ (or larger) models, and their approach can therefore only capture limited textural information.

To avoid the above problems, image restoration is typically performed using gradient-ascent, thereby eliminating the need to deal with many discrete gray-levels, and avoiding expensive sampling \cite{roth05fields}. While gradient-based approaches are generally considered to be fast, they may still require several thousand iterations in order to converge, and even then will converge only to a \emph{local} optimum.

In this paper, we propose a method that gives us the best of both worlds: we manage to render belief-propagation practical using a high-order ($2\times 2$) model, and use it for the task of image inpainting. By using a nonparametric prior\footnote{The term `nonparametric' is something of a misnomer, since the prior is actually approximated using a mixture of Gaussians. This term was originally used in \cite{sudderth02nonparametric}, and is maintained here for consistency.}, we avoid the need to discretize images, resulting in much smaller messages being passed between cliques. Our experiments show that belief-propagation techniques are able to produce competitive results after only a single iteration, rendering them faster than many gradient-based approaches, while retaining similar visual quality of the restoration.

\section{Background}
\label{sec:background}

In this section, we define the Markov random field (MRF) image prior to be used in our model. Although we shall not present any significant results in terms of \emph{learning} the prior, we have nevertheless made a number of modifications to the `standard' image prior in order to render inference tractable.

\subsection{The `Field of Experts' Image Prior}

The Hammersley-Clifford theorem states that the joint probability distribution of a Markov random field with clique set $\mathcal C$ (assuming maximal cliques) is given by
\begin{equation}
p(\mathbf{x}) = \frac{1}{Z}\prod_{c\in\mathcal C} \phi_c(\mathbf{x}_c)
\end{equation}
(where $\mathbf{x}_c$ is the set of variables in $\mathbf{x}$ belonging to the $c^{th}$ clique; $Z$ is a normalization constant) \cite{Geman84stochastic}. When dealing with images, the $\phi_c$s are often assumed to be homogeneous \cite{BurMoo87}, meaning that the prior can be defined entirely in terms of a single potential function, $\phi$. In the Field of Experts model \cite{roth05fields}, this potential function is assumed to take the form of a \emph{product of experts} \cite{Hin99}, in which each `expert' is the response of the image patch \{$\mathbf{x}_c$\} to a particular filter \{$J_f$\}. That is, the potential function takes the form
\begin{equation}
\phi(\mathbf{x}_c; J, \alpha) = \prod_{f=1}^F \phi'_f(\mathbf{x}_c, J_f, \alpha_f)
\label{eq:pot1}
\end{equation}
(where the $\alpha_f$'s are simply weighting coefficients controlling the importances of the filters). Specifically, each expert is assumed to take the form of a Student's T-distribution, namely
\begin{equation}
\phi'_f(\mathbf{x}_c; J_f, \alpha_f) = (1 + \frac{1}{2}\left\langle J_f, \mathbf{x}_c \right\rangle^2)^{-\alpha_f}.
\label{eq:pot2}
\end{equation}
Although \cite{roth05fields} use contrastive divergence learning to select the filters and alphas, it has been shown that the filters can more easily be selected using principal component analysis (PCA) \cite{mca06}. This leaves only the problem of learning the alphas, which we shall deal with in section \ref{sec:approx}.

\subsection{Belief-Propagation}
\label{sec:bprop}

Inference in the MRF setting can be formulated as a message passing problem. Two common message passing algorithms exist, namely the \emph{junction-tree algorithm}, and \emph{loopy belief-propagation} \cite{thegdl}. In our case, which algorithm should be applied depends upon the `shape' of the region being inpainted. We will give only a brief overview of these algorithms in order to explain why it is infeasible to apply them directly when using the above prior. A more complete specification is given in \cite{thegdl}; similar ideas are also used in an image inpainting setting in \cite{levin03}.

Belief-propagation algorithms work by having cliques pass `messages' to other cliques which share one or more nodes in common. If we denote by $S_{i,j}$ the intersection of the two cliques $\mathbf{x}_i$ and $\mathbf{x}_j$, and denote by $\Gamma\!_{\mathbf{x}_i}$ the neighbors of $\mathbf{x}_i$ (i.e.\ those cliques which share one or more nodes with $\mathbf{x}_i$), then the message, $M_{i\rightarrow j}$, sent from $\mathbf{x}_i$ to $\mathbf{x}_j$ is given by
\begin{equation}
M_{i\rightarrow j}(S_{i,j}) = \sum_{\mathbf{x}_i\backslash \mathbf{x}_j}\!\phi(\mathbf{x}_i) \hspace{-5mm} \prod_{\mathbf{x}_k \in (\Gamma\!_{\mathbf{x}_i}\!\backslash \lbrace \mathbf{x}_j \rbrace)} \hspace{-6mm} M_{k\rightarrow i}(S_{k,i}).
\label{eq:message}
\end{equation}
That is, the outgoing message from $\mathbf{x}_i$ to $\mathbf{x}_j$ is defined as the product of the local potential \{$\phi(\mathbf{x}_i)$\} with the incoming messages from all neighbors \emph{except} $\mathbf{x}_j$, marginalized over the variables in $\mathbf{x}_i$ but not in $\mathbf{x}_j$. Once all messages have been sent, the final distribution of $\mathbf{x}_i$ \{$D_i(\mathbf{x}_i)$\} is given by
\begin{equation}
D_i(\mathbf{x}_i) = \phi(\mathbf{x}_i) \!\! \prod_{\mathbf{x}_k \in \Gamma\!_{\mathbf{x}_i}} \!\! M_{k\rightarrow i}(S_{k,i}).
\label{eq:messageproduct}
\end{equation}
Even when using the $2\times 2$ model, evaluating $\phi(\mathbf{x}_i)$ requires us to consider $256^4$ possible gray-level combinations. Although it may be possible to approximate the marginal being computed in equation (\ref{eq:message}) without computing $\phi(\mathbf{x}_i)$ explicitly \cite{lanroth}, the message itself still contains $256^2$ elements. This problem is dealt with in \cite{lanroth} by using a \emph{factor-graph}  \cite{kschischang01factor}, which requires only that one dimensional marginals are computed; however the running time of their method is still linear in the number of gray-levels, in addition to the fact that the factor-graph fails to fully capture the conditional independencies implied by the model.

As a result, we seek $\phi$ in such a form that the sum in equation (\ref{eq:message}) may be replaced by an integral. In \cite{sudderth02nonparametric}, the authors defined such a model in which the potential function takes the form of a Gaussian mixture, that is, with $\phi$ taking the form
\begin{equation}
\phi(\mathbf{x}_c) = \sum_{i=1}^N \beta_ie^{(\mathbf{x}_c - \mu_i)^T\Sigma_i^{-1}(\mathbf{x}_c - \mu_i)}
\label{eq:mog}
\end{equation}
(this is sometimes known as a \emph{Gaussian random field} \cite{bishop06,GMRFbook}). Unfortunately, the method they use to learn these mixtures appears to be applicable only to low-order models (the largest mixture models they learn are 3-dimensional).

In the remainder of this paper, we will show that the experts \{$\phi'_f$s\} can be approximated as a Gaussian mixture, resulting in a high-order model which closely matches the one given in equations (\ref{eq:pot1}) and (\ref{eq:pot2}). We will describe belief-propagation in this setting, and show how this approach can be used for fast image inpainting.

\section{Approximating the Prior}
\label{sec:approx}

\begin{figure}[t]
\begin{center}
\includegraphics[width=0.3\columnwidth, height=0.25\columnwidth]{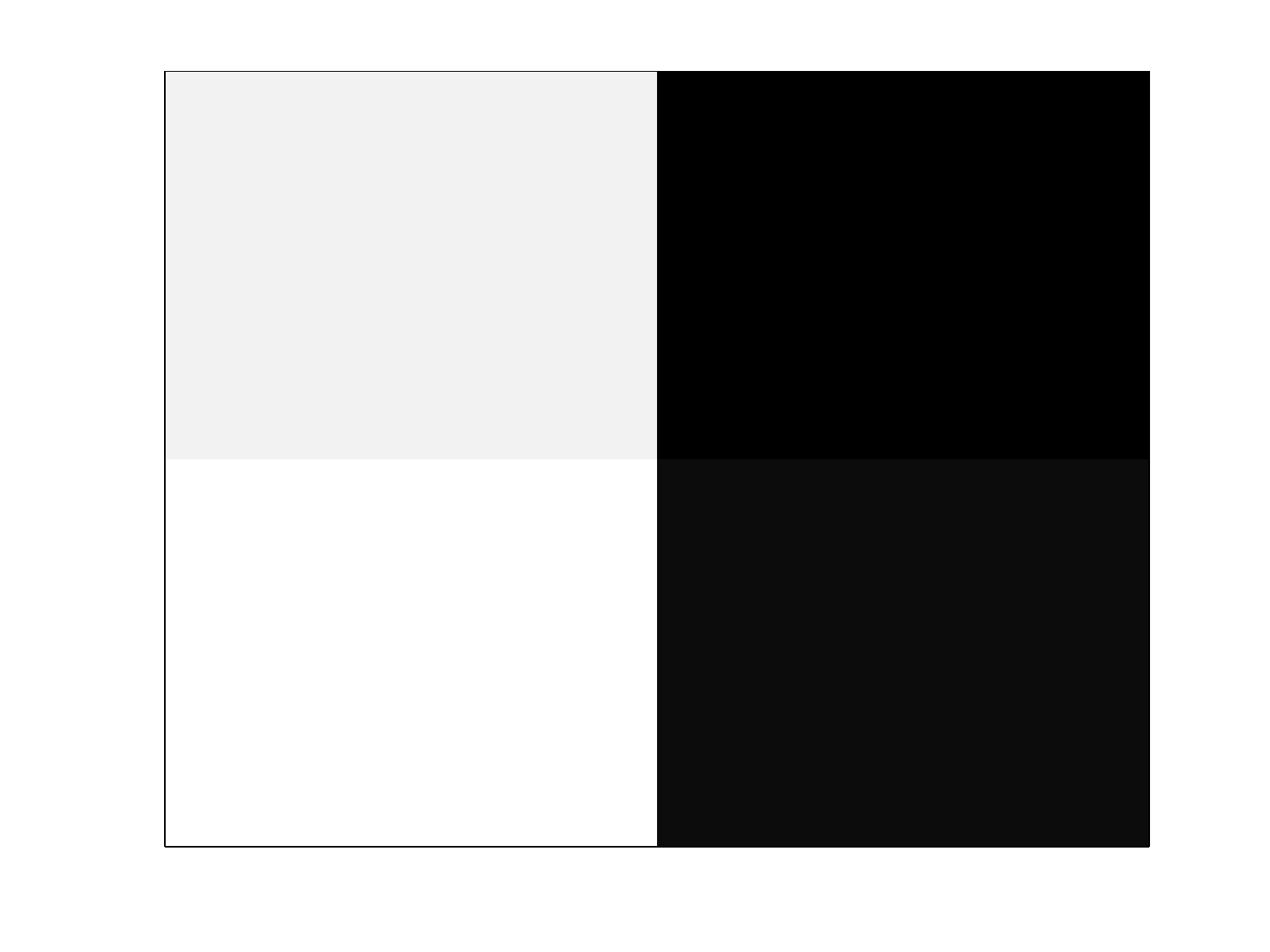}
\includegraphics[width=0.3\columnwidth, height=0.25\columnwidth]{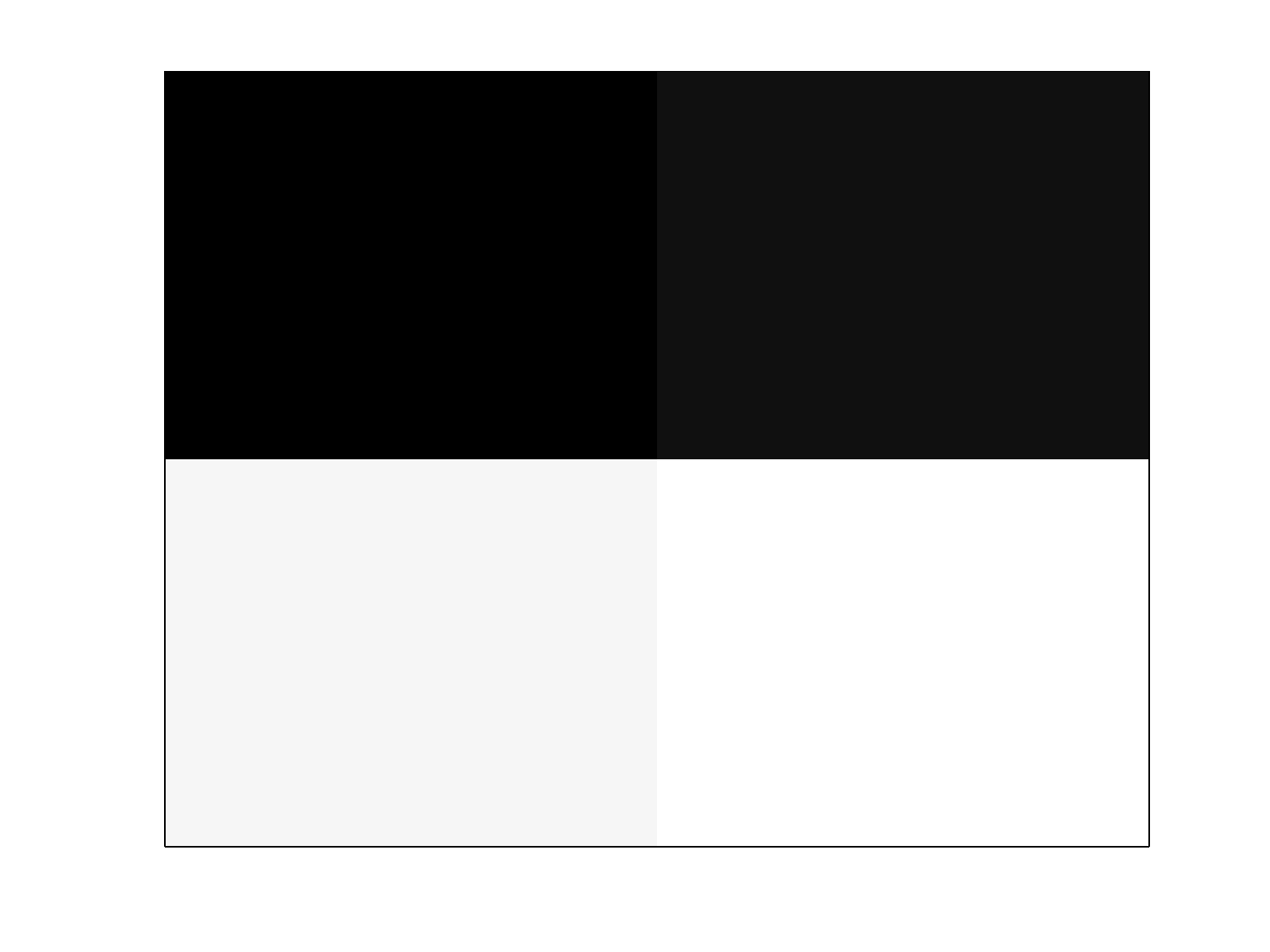}
\includegraphics[width=0.3\columnwidth, height=0.25\columnwidth]{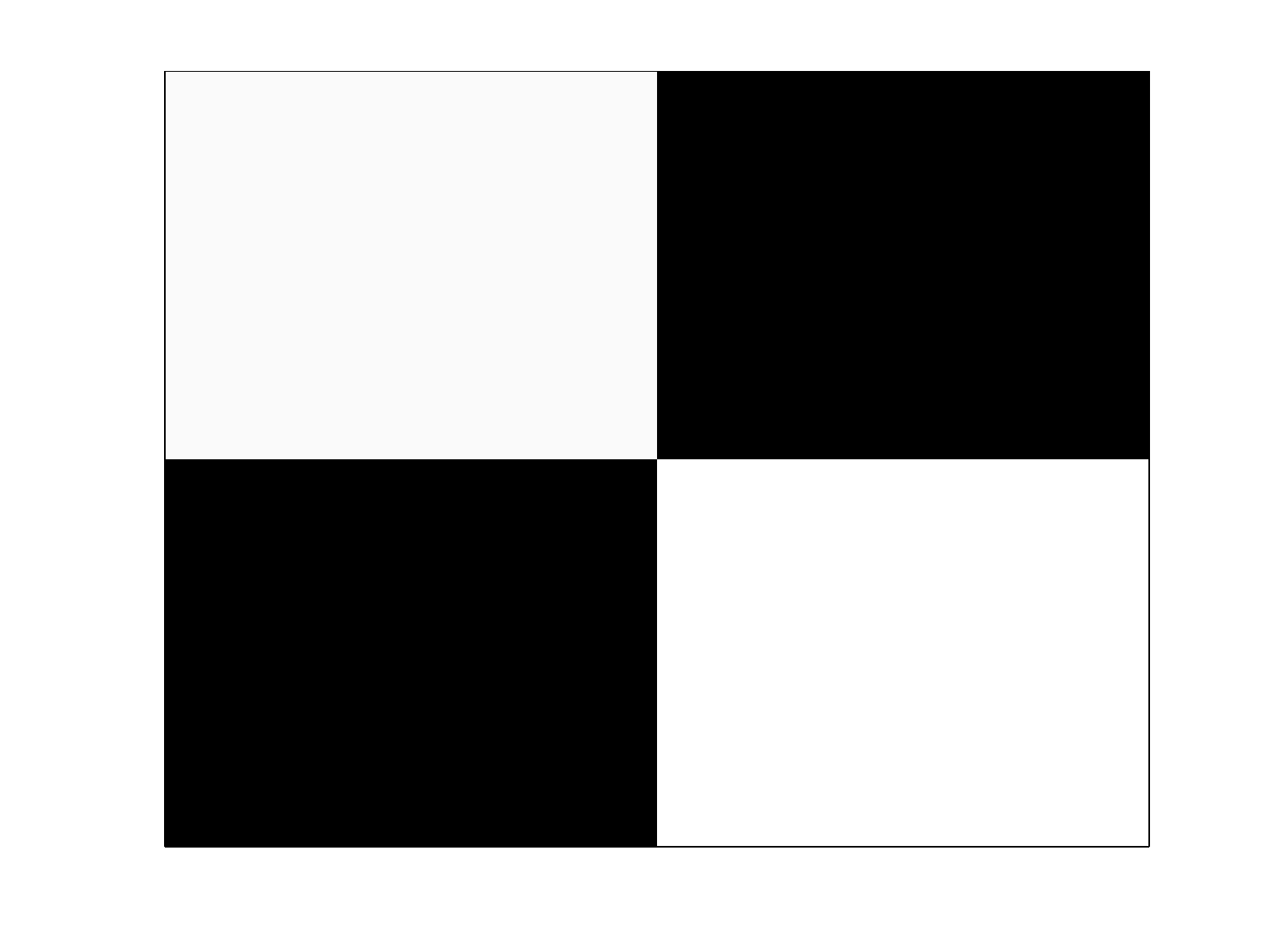}
\end{center}
\caption{The three filters used in our $2\times 2$ model.}
\label{fig:filters}
\end{figure}

In \cite{mca06}, the authors showed that the filters \{$J_f$s\} in equation (\ref{eq:pot1}) can be learned by performing a PCA on a collection of natural image patches. Here we follow this idea. To learn our filters (for a $2\times 2$ model), we randomly cropped 50,000 $2\times 2$ patches from images in the Berkeley Segmentation Database \cite{Mar01}, and used their principal components as our filters. It was found in \cite{roth05fields} that the first component of such a PCA always corresponds to a uniform gray patch, which should be ignored in order to obtain a model invariant to intensity. hence we only use the \emph{last three} filters for our $2\times 2$ model. The resulting patches appear to make sense visually, and are shown in figure \ref{fig:filters}. This technique can also be used to learn $3\times 3$ or $5\times 5$ models (resulting in 8 or 24 filters, respectively), which are not shown.

This model requires also that we learn the `importances' $\{\alpha_f\}$ of each filter. From equation (\ref{eq:pot2}), it can be seen that the $\alpha_f$s simply control the shape (or `peakedness') of the Student's T-distribution. Rather than try to learn the $\alpha_f$s explicitly, we will learn them implicitly through our approximation.

\begin{figure}[t]
\begin{center}
\includegraphics[width=\columnwidth]{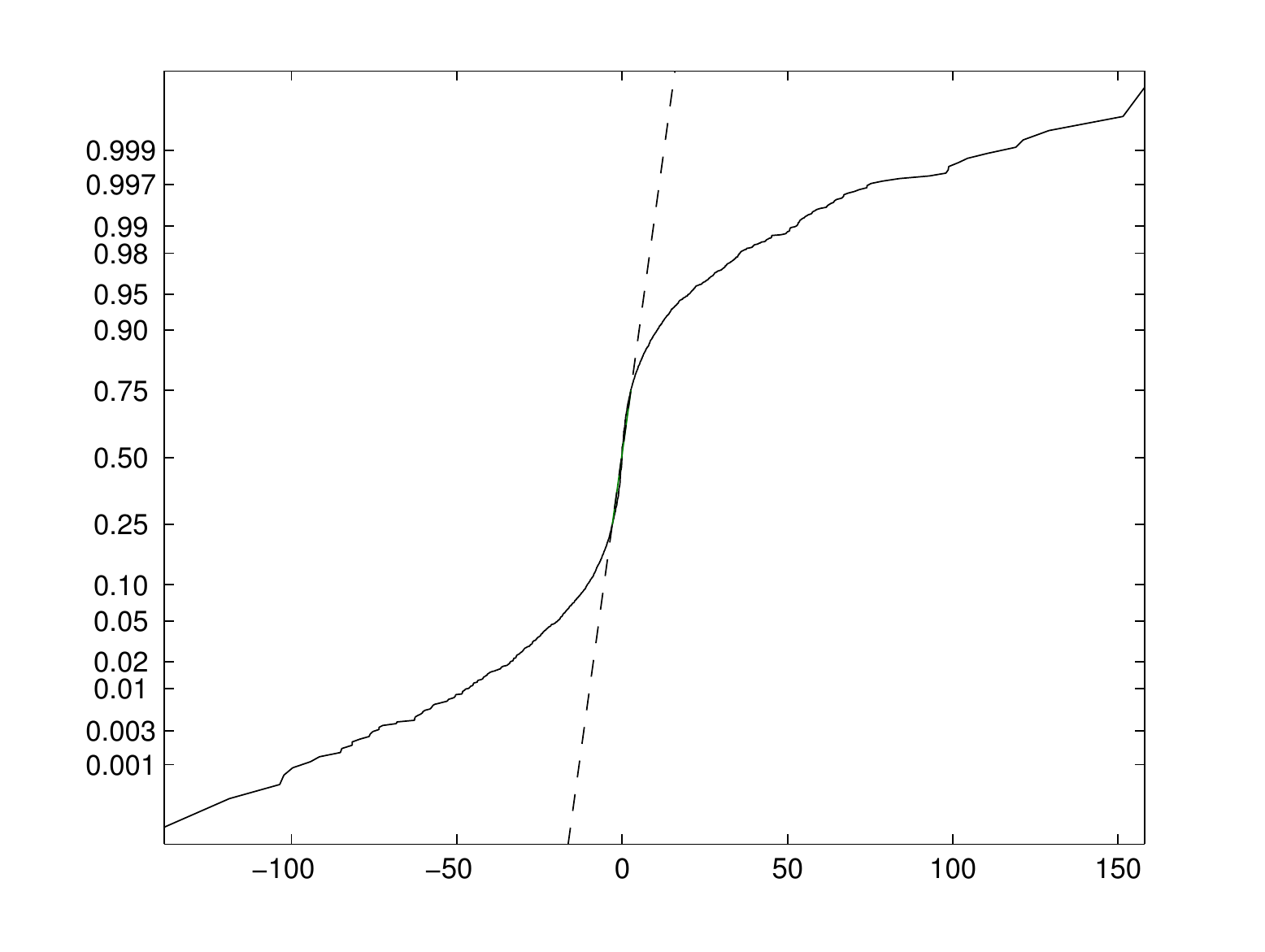}
\end{center}
\caption{Solid line: A normal probability plot of the inner products (horizontal axis) against their corresponding normal probabilities (vertical axis). This plot shows that the inner products are more heavily tailed than would be suggested by a normal distribution (dotted line).}
\label{fig:normplot}
\end{figure}

In order to approximate the experts $\{\phi'_f\}$, we first calculated the inner products $\{\left\langle J_f, \mathbf{x}_c \right\rangle\}$ for a random selection of 5,000 image patches (again cropped from the Berkeley segmentation database \cite{Mar01}). A normal probability plot of the data (against the first filter) is shown in figure \ref{fig:normplot} -- this plot reveals that the data is more heavily tailed than would be suggested by a normal distribution, indicating that the Student's T-distribution may indeed be valid. However, rather than assume that this data is generated according to a Student's T-distribution, we simply tried to approximate it directly using a mixture of Gaussians.

In order to estimate the distribution governing this data, we used the expectation-maximization (EM) algorithm \cite{dlr77}, assuming that the set of inner products for each filter was generated by a mixture of three Gaussians. All of our parameters to be learned \{$\Theta = (\beta, \mu, \sigma)$\} were initialized by using a K-means clustering \cite{MacQueen67} on the original inner products. We used this approach to learn a separate mixture model for each expert. This algorithm produces an approximation of the form
\begin{equation}
\phi'_f(\mathbf{x}_c; \Theta, J) \simeq \sum_{i=1}^3\beta_{f,i}\exp\left( {\frac{(\left\langle J_f, \mathbf{x}_c \right\rangle - \mu_{f,i})^2}{2\sigma_{f,i}^2}} \right).
\label{eq:approximation}
\end{equation}
The alpha terms are no longer relevant -- the `shape' of the distribution is implicitly controlled by the other parameters. However, the expression in equation (\ref{eq:approximation}) is not yet in the same form as equation (\ref{eq:mog}). Hence we need to solve the system
\begin{equation}
\exp\left( {\frac{(\left\langle J, \mathbf{x} \right\rangle - \mu)^2}{2\sigma^2}} \right) = \exp\left( (\mathbf{x} - \underline{\mu})^T \Sigma^{-1} (\mathbf{x} - \underline{\mu}) \right).
\end{equation}
That is, we are trying to solve for $\Sigma^{-1}$ (a matrix) and $\underline{\mu}$ (a vector), in terms of $J$ (a vector) and $\mu$ (a scalar). It is not difficult to see that the only solution for $\Sigma^{-1}$ is
\begin{equation}
\Sigma^{-1} = \frac{1}{2\sigma^2}J\cdot J^T = 
\frac{1}{2\sigma^2} \left[ \begin{array}{cccc}
                      J_1^2 & J_1J_2 & \cdots & J_1J_n\\
                      J_2J_1 & J_2^2 & & J_2J_n\\
                      \vdots & & \ddots & \vdots\\
                      J_nJ_1 & J_nJ_2& \cdots & J_n^2
                    \end{array} \right]
\label{eq:covinv}
\end{equation}
(where $n$ is the size of the filter $J$ -- in our case, $n~=~4$). Alternately, there are infinite solutions for $\underline{\mu}$. One obvious solution is
\begin{equation}
\underline{\mu} = \left( \begin{array}{c}
                          \frac{\mu}{\sum_{i=1}^nJ_i}\\
                          \vdots\\
                          \frac{\mu}{\sum_{i=1}^nJ_i}
                          \end{array}
                  \right).
\end{equation}
However, we found for all of our filters that $\sum_{i=1}^nJ_i~\!\!\simeq~\!\!0$, meaning that this solution would be highly unstable. A more stable solution (which we used) is given by
\begin{equation}
\underline{\mu} = \left( \begin{array}{c}
                          \mu/J_1\\
                          0\\
                          \vdots\\
                          0
                         \end{array} \right).
\end{equation}

Our potential function is now of the form
\begin{equation}
\phi(\mathbf{x}_c; \Theta, J) = \prod_{f=1}^3\sum_{i=1}^3\beta_{f,i}\exp\left( {\frac{(\left\langle J_f, \mathbf{x}_c \right\rangle - \mu_{f,i})^2}{2\sigma_{f,i}^2}} \right).
\end{equation}
In order to expand the above product, we use the following result about the product of Gaussian distributions \cite{ahrendt05}: given $K$ Gaussians (with means $\mu_1 \ldots \mu_K$, and covariances $\Sigma_1 \ldots \Sigma_K$), the covariance of the product \{$\Sigma'$\} is given by
\begin{equation}
\Sigma' = (\sum_{i=1}^K\Sigma_i^{-1})^{-1}
\label{eq:prod1}
\end{equation}
(although each $\Sigma_i^{-1}$ is singular in our case, their sum is not). The mean of the product \{$\mu'$\} is given by
\begin{equation}
\mu' = \Sigma'(\sum_{i=1}^K\Sigma_i^{-1}\mu_i).
\label{eq:prod2}
\end{equation}
The corresponding beta term for the product is just $\beta' = \prod_{i=1}^K\beta_k$. In our case this results in a final approximation which is a mixture of $3^3 = 27$ Gaussians.

\section{Inference}
\label{sec:inference}

In order to perform belief-propagation, we must first be able to express equations (\ref{eq:message}) and (\ref{eq:messageproduct}) in terms of the Gaussian mixtures we have defined. In our setting, the sum in equation (\ref{eq:message}) becomes an integral, resulting in the new equation
\begin{equation}
M_{i\rightarrow j}(S_{i,j}) = \int_{\mathbf{x}_i\backslash \mathbf{x}_j}\!\!\!\phi(\mathbf{x}_i) \hspace{-5mm} \prod_{\mathbf{x}_k \in (\Gamma\!_{\mathbf{x}_i}\!\backslash \lbrace \mathbf{x}_j \rbrace)} \hspace{-6mm} M_{k\rightarrow i}(S_{k,i}).
\label{eq:newmessage}
\end{equation}
We have already suggested how to perform the above multiplication in equations (\ref{eq:prod1}) and (\ref{eq:prod2}). The only difference in this case is that the mixtures for each message may contain fewer variables (and smaller covariance matrices) than the local distribution \{$\phi(\mathbf{x}_i)$\}. In such a case, the inverse covariance matrices for each message \{$\Sigma^{-1}$s\} are simply assumed to be zero in all missing variables.

To compute the marginal distribution of a Gaussian mixture with mean $\mu$ and covariance matrix $\Sigma$ (i.e.\ the integral in equation (\ref{eq:newmessage})), we simply take the elements of $\mu$ and $\Sigma$ corresponding to the variables whose marginals we want. The importances for each Gaussian in the mixture remain the same.

Of course, when we compute the products in equation (\ref{eq:newmessage}), we produce a model with an exponentially increasing number of Gaussians. As a simple solution to this problem, we restrict the maximum number of Gaussians to a certain limit (see section \ref{sec:results}), by including only those with the highest importances.

When solving an inpainting problem, we only wish to treat some of the variables in each clique as unknowns (for example, the `scratched' sections). Hence the potential function for these cliques should be conditioned upon the `observed' regions of the image. Suppose that for a clique $c$ we have unknowns $\mathbf{x}_{(u)}$, and observed variables $\mathbf{x}_{(o)}$ (i.e.\ $\mathbf{x}_c = (\mathbf{x}_{(u)}^T;\mathbf{x}_{(o)}^T)^T$). Then we may partition the mean and covariance matrix (for a particular Gaussian in the mixture) as
\begin{equation}
\mu = \left( \begin{array}{c}
               \mu_{(u)}\\
               \mu_{(o)}
             \end{array} \right)
\end{equation}
and
\begin{equation}
\Sigma = \left[ \begin{array}{cc}
                  \Sigma_{(u,u)} & \Sigma_{(u,o)}\\
                  \Sigma_{(o,u)} & \Sigma_{(o,o)}
                \end{array} \right].
\end{equation}
The mean of the conditional distribution \{$\mu_{(u;o)}$\} is now given by
\begin{equation}
\mu_{(u;o)} = \mu_{(u)} + \Sigma_{(u,o)}\Sigma_{(o,o)}^{-1}(\mathbf{x}_{(o)} - \mu_{(o)}),
\end{equation}
and the covariance matrix \{$\Sigma_{(u;o)}$\} is given by
\begin{equation}
\Sigma_{(u;o)} = \Sigma_{(u,u)} - \Sigma_{(u,o)}\Sigma_{(o,o)}^{-1}\Sigma_{(u,o)}^T.
\end{equation}

Finally, once all messages have been propagated, we are able to compute the marginal distribution for a given node (or pixel, belonging to clique $c$) by marginalizing $D_c(\mathbf{x}_c)$ (equation (\ref{eq:messageproduct})) in terms of that node. In order to estimate the `most likely' configuration for this pixel, we simply consider each of the 256 possible gray-levels.\footnote{Although this final step may appear to make the running time of our solution linear in the number of gray-levels, it should be noted that this step needs to be performed only once, after the final iteration. It should also be noted that this estimate only requires us to measure the response of a one-dimensional Gaussian, which is inexpensive. More sophisticated mode-finding techniques exist \cite{miguel00mode}, which we considered to be unnecessary in this case. Finally, note that this step is not required when our mixture contains only a single Gaussian, in which case we simply select the mean.}

\subsection{Propagation Methods}
\label{sec:propmethods}

As we mentioned in section \ref{sec:bprop}, the two propagation techniques we will deal with are the junction-tree algorithm and loopy belief-propagation. Although we will not cover these in great detail (see \cite{thegdl} for a more complete exposition), we will explain the differences between the two in terms of image inpainting.

Both algorithms work by passing messages between those cliques with non-empty intersection. However, when using the junction-tree algorithm, we connect only enough cliques to form a maximal spanning tree. Now, suppose that two cliques $c_a$ and $c_b$ have intersection $S_{a,b}$. If each clique along the path between them also contains $S_{a,b}$, we say that this spanning tree obeys the `junction-tree property'. If this property holds, it can be proven that exact inference is possible (subject only to the approximations used by our Gaussian model), and requires that messages be passed only for a single iteration \cite{thegdl}. Technically, the graphs which obey this property are the so-called\emph{triangulated} or \emph{chordal} graphs. The tractability of exact inference in these graphs depends on their \emph{tree-width}: graphs that are more `tree-like' are better suited to efficient exact inference. See \cite{bishop06} for details.

If this property \emph{doesn't} hold, then we may resort to using loopy belief-propagation, in which case we simply connect \emph{all} cliques with non-empty intersection. There is no longer any message passing order for which equation (\ref{eq:newmessage}) is well defined (i.e.\ we must have a criterion to initialize some messages, and the common choice is to assume they have a uniform distribution), meaning that messages must be passed for many iterations in the hope that they will converge.

\begin{figure}[t]
\begin{center}
\includegraphics[width=0.28\columnwidth]{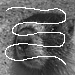}
\hspace{0.02\columnwidth}
\includegraphics[width=0.28\columnwidth]{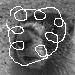}
\hspace{0.02\columnwidth}
\includegraphics[width=0.28\columnwidth]{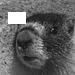}
\end{center}
\caption{The graph formed from the white pixels in the left image forms a junction-tree (assuming a $2\times 2$ model). The graphs formed from the white pixels in the other two images do not.}
\label{fig:animal}
\end{figure}

Figure \ref{fig:animal} shows an inpainting problem for which a junction-tree exists, and two problems for which one does not (assuming $2\times 2$-pixel cliques). Since the regions being inpainted are usually thin lines (or `scratches'), we may often observe graphs which do in fact obey the junction-tree property in practice.

Fortunately, we found that even in those cases where no junction-tree existed, loopy belief-propagation tended to converge in very few iterations. Although there are few theoretical results to justify this behavior, loopy-belief propagation typically converges quickly in those cases where the graph \emph{almost} forms a tree (as is usually the case for the regions being inpainted).

\section{Experimental Results}
\label{sec:results}

In order to perform image inpainting, we used a high-level (Python) implementation of the junction-tree algorithm and loopy belief-propagation, which is capable of constructing Markov random fields with any topology. Despite being written in a high-level language, our implementation is able to inpaint images within a reasonably short period of time. Since it is difficult to assess the quality of our results visually, we have reported both the peak signal-to-noise ratio (PSNR), and the structured similarity (SSIM) \cite{wang04}.

\begin{figure}[t]
\begin{center}
\includegraphics[width=0.45\columnwidth]{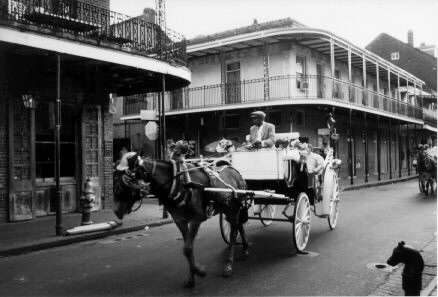}\hspace{1mm}\includegraphics[width=0.45\columnwidth]{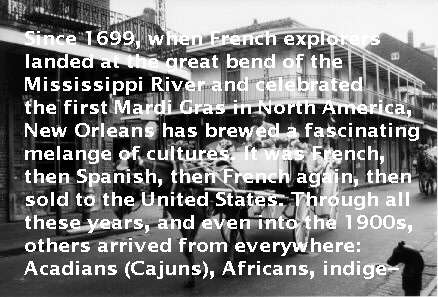}\\
\vspace{0.5mm}
\includegraphics[width=0.45\columnwidth]{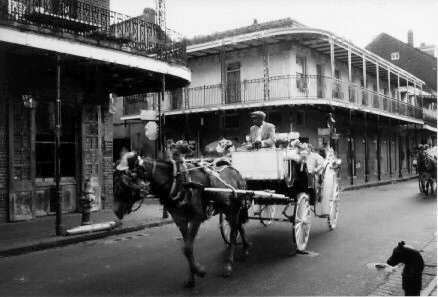}\hspace{1mm}\includegraphics[width=0.45\columnwidth]{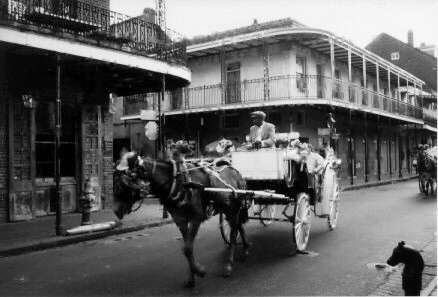}\\
\vspace{-2.5mm}
\rule{0.94\columnwidth}{0.2mm}\\
\vspace{1mm}
\includegraphics[width=0.45\columnwidth]{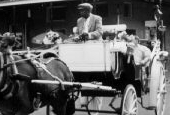}\hspace{1mm}\includegraphics[width=0.45\columnwidth]{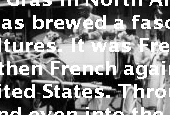}\\
\vspace{0.5mm}
\includegraphics[width=0.45\columnwidth]{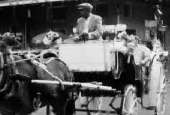}\hspace{1mm}\includegraphics[width=0.45\columnwidth]{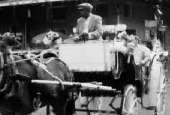}
\end{center}
\caption{Above, top-left to bottom-right: the original image; the image containing the text to be removed; inpainting after a single iteration, using a single Gaussian (PSNR = 22.74, SSIM = 0.962); inpainting after two iterations (PSNR = 22.82, SSIM = 0.962). Below: close-ups of all images.}
\label{fig:text}
\end{figure}

We were not able to directly compare our PSNR results to those in \cite{roth05fields}, since they only presented $3\times 3$ and $5\times 5$ models. While it is certainly true that their $3\times 3$ model produces a \emph{much} higher PSNR than our technique (e.g.\ a PSNR of $\sim\!\!31.4$ for the image in figure \ref{fig:text}), its execution time is simply impractical. Fortunately, it is still possible to measure approximate execution times of a $2\times 2$ model using their approach, even without reporting PSNRs. These results are presented in the next section.

While it is true that the difference between the two models being compared makes meaningful comparison difficult, it is most important to note that there is little \emph{visual} difference between the two models. In the next section, we will show that our $2\times 2$ model is faster than a similar model using gradient-ascent -- it is the combination of these two results which we believe makes our technique viable.

Figure \ref{fig:text} shows a corrupted image from which we want to remove the text. The image has been inpainted using a model containing only a single Gaussian (although the learned mixtures contained three Gaussians -- see below). After a single iteration, most of the text has been removed, and after two iterations it is almost completely gone. Although the current state-of-the-art inpainting techniques produce superior results in terms of PSNR \cite{roth05fields}, they give similar visual results and take several thousand iterations to converge, compared to ours which takes only two (no further improvement was observed after a third iteration).

\begin{figure}[t]
\begin{center}
\includegraphics[width=0.45\columnwidth]{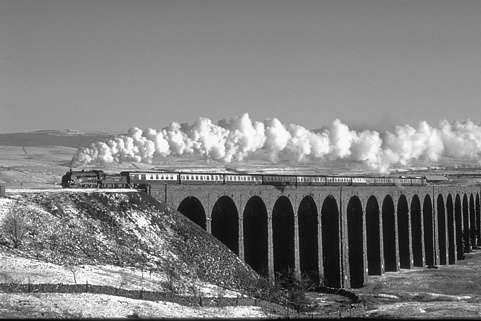}\hspace{1mm}\includegraphics[width=0.45\columnwidth]{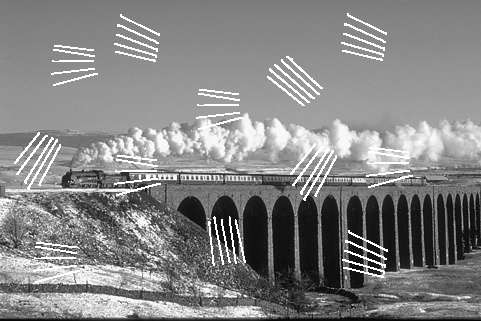}\\
\vspace{-2.5mm}
\rule{0.94\columnwidth}{0.2mm}\\
\vspace{1mm}
\includegraphics[width=0.45\columnwidth]{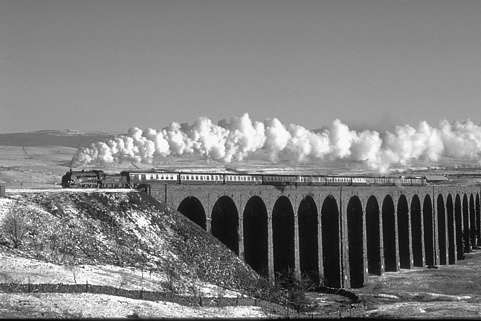}\hspace{1mm}\includegraphics[width=0.45\columnwidth]{output3_31.jpg}\\
\vspace{0.5mm}
\includegraphics[width=0.45\columnwidth]{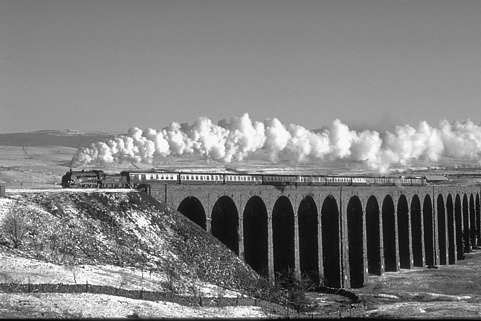}\hspace{1mm}\includegraphics[width=0.45\columnwidth]{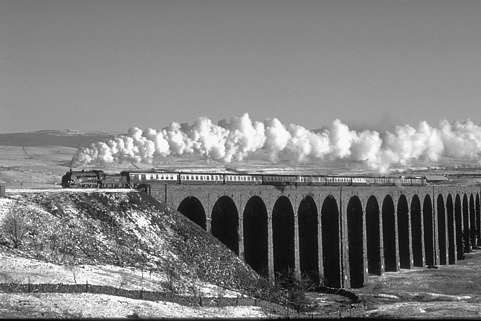}
\end{center}
\caption{Above: the original image; the corrupted image containing `scratches'. Below, top-left to bottom-right: mixture contains 1 Gaussian; mixture contains 3 Gaussians, propagation is performed with 1 Gaussian; propagation is performed with 3 Gaussians; propagation is performed with 9 Gaussians. All results are shown using the $2\times 2$ model, after three iterations. See table \ref{tab:results} for more detail.}
\label{fig:results}
\end{figure}

Figure \ref{fig:results} compares models of various sizes, varying both the number of Gaussians used to approximate each mixture, as well as the maximum number of Gaussians allowed during the inference stage. The same results are summarized in table \ref{tab:results}.

\begin{table}
\begin{center}
\begin{tabular}{|l|l|l|l|l|}
\hline
Gaussians & Max & Iter. & PSNR & SSIM\\
\hline
1 & 1 & 1 & 22.57 & 0.927\\
  &   & 2 & 22.67 & 0.928\\
  &   & 3 & 22.68 & 0.928\\
\hline
3 & 1 & 1 & 22.81 & 0.927\\
  &   & 2 & 22.87 & 0.928\\
  &   & 3 & 22.88 & 0.928\\
\hline
3 & 3 & 1 & 22.82 & 0.927\\
  &   & 2 & 22.87 & 0.928\\
  &   & 3 & 22.88 & 0.928\\
\hline
3 & 9 & 1 & 22.80 & 0.927\\
  &   & 2 & 22.86 & 0.928\\
  &   & 3 & 22.87 & 0.928\\
\hline
\end{tabular}
\end{center}
\caption{Comparison of inpainting performance for several models. Here we vary the number of Gaussians used to compute the initial mixture, as well as the maximum number of Gaussians allowed during propagation.}
\label{tab:results}
\end{table}

The top-right image in figure \ref{fig:results} was produced using a model in which each expert was approximated using three Gaussians, yet only one Gaussian was allowed during propagation. In contrast, the model used to produce the top-left image was approximated using only a \emph{single} Gaussian. Interestingly, the former model actually outperformed the latter in this experiment. While this result may seem surprising, it may be explainable as follows: in the single-Gaussian model, the standard deviation is overestimated in order to compensate for the high kurtosis of the training data \cite{student1908}. However, in the model containing three Gaussians, the most significant Gaussian (i.e.\ the Gaussian with the highest $\beta$ term) captures only the most of the `important' information about the distribution, and ignoring the other two is not very harmful.

Furthermore, given that increasing the maximum number of Gaussians allowed during propagation does not seem to significantly improve inpainting performance, we suggest that this single-Gaussian model may be the most practical. Even after only a single iteration, the results are visually pleasing.

Finally, although we mentioned that it is easy to \emph{learn} $3\times 3$ or larger models using the proposed method, such models were found to be impractical in an inference setting. For example, a $3\times 3$ model would have a total of $8$ filters, resulting in a mixture model with $3^8$ Gaussians. Although this problem may again be addressed by simply including only the most important Gaussians, this results in a very high error in the approximation; we found that this model did not outperform the $2\times 2$ version in practice.

\subsection{Execution Times}

Unfortunately, it proved very difficult to compare the execution times of our model with existing gradient-ascent techniques. For example, the inpainting algorithm used in \cite{roth05fields} computes the gradient for all pixels using a 2-dimensional matrix convolution over the \emph{entire} image, and then selects only the region corresponding to the inpainting mask. While this results in very fast performance when a reasonable proportion of an image is being inpainted, it results in very slow performance when the inpainting region is very sparse (as is often the case with scratches). It is easy to produce results which favor either algorithm, but such a comparison will likely be unfair.

\begin{figure}[t]
\begin{center}
\includegraphics[width=0.45\columnwidth]{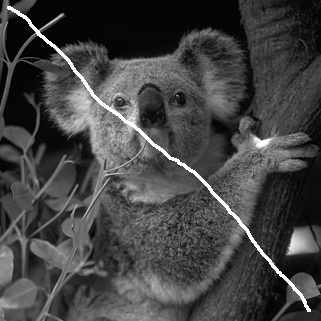}\hspace{1mm}\includegraphics[width=0.45\columnwidth]{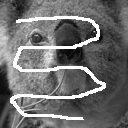}
\end{center}
\caption{Two equally large regions to be inpainted, in two differently sized images.}
\label{fig:koalas}
\end{figure}

To make explicit this difficulty, consider the images in figure \ref{fig:koalas}. The image on the left is significantly larger than the image on the right, yet the corrupted regions are of the same size ($\sim\!\!1500$ pixels). As a result, our algorithm exhibited the same running time on both images, whereas the gradient-ascent algorithm from \cite{roth05fields} was approximately 6 times slower on the larger image.

As a more representative example, when inpainting the image in figure \ref{fig:results} (using a single Gaussian), the first iteration took $\sim\!\!33.6$ seconds on our test machine. The second iteration took $\sim\!\!39.0$ (as did subsequent iterations -- the first is slightly faster due to many messages being empty at this stage). The running time of this algorithm increases linearly with the number of Gaussians (for example, when using three Gaussians, the first iteration took $\sim\!\!87.0$ seconds).

Alternately, a single iteration of inpainting using the gradient-ascent algorithm from \cite{roth05fields} took $\sim\!\!0.1$ seconds (using a $2\times 2$ model). However, their code was run for 2,500 iterations, meaning that our code is still in the order of 2 to 3 times faster. This is a pleasing result, given that we used a high-level language for our implementation.

However, in an attempt to provide a more `fair' comparison, we have tried to analyze the computations required by both algorithms. It can be seen from the equations presented in section \ref{sec:inference} that our algorithm consists (almost) entirely of matrix multiplications and inverses.\footnote{Other operations, such as additions and permutations are typically much faster than these.} Although it is very difficult to express exactly the number of such operations required by our algorithm in general, we have calculated this number for a specific case.

The corrupted image in figure \ref{fig:results} requires us to inpaint a total of 5829 pixels. The number of operations required by our algorithm to inpaint this image (during the \emph{second} iteration) is shown in table \ref{tab:runtime}.

Alternately, the gradient-ascent approach in \cite{roth05fields} is dominated by the time taken to compute the inner products in (the derivative of the logarithm of) equation (\ref{eq:pot2}). Each pixel is contained by four cliques, and we must compute the inner product against each of our three filters. Therefore we must compute a total of $4\times 3\times 5829 = 69948$ inner products per iteration.

\begin{table}
\begin{center}
\begin{tabular}{|l|l|l|}
\hline
$n$ & Multiplications & Inverses\\
\hline
1 & 14800 & 44648\\
2 & 42072 & 37386\\
3 & 25760 & 12880\\
4 & 43308 & 21654\\
\hline
\end{tabular}
\end{center}
\caption{Number of operations required by our algorithm. Multiplications are of $(n\times n)\times(n\times 1)$ matrices, and inverses are of $(n\times n)$ matrices, for various values of $n$.}
\label{tab:runtime}
\end{table}

As a simple experiment, we timed these operations in Matlab (using random matrices and vectors). We found that computing 69948 inner products was approximately 10 times faster than computing the matrix operations shown in table \ref{tab:runtime}. This leads us to believe that a low-level implementation of our belief-propagation algorithms may be significantly faster even than the results we have shown.

\section{Discussion}

Our results have shown than even a $2\times 2$ model is able to produce very satisfactory inpainting performance. We believe that even this small model is able to capture much of the important information about natural images. While higher-order models exist \cite{roth05fields}, the improvements appear to be quite incremental, despite a significant increase in their execution time. While it is certainly the case that our results fall short of the state-of-the-art in terms of PSNR, the differences are difficult to distinguish visually. It is therefore pleasing that we are able to produce competitive results within only a short period of time.

We have not yet fully explored the possibility of using the junction-tree algorithm to inpaint images. Unfortunately, determining whether a graph obeys the junction-tree property (see section \ref{sec:propmethods}) is very expensive, meaning we simply used loopy belief-propagation in all cases, without even performing this test. However, there are many cases in which we can be \emph{sure} that a junction-tree exists -- for example, if the inpainting region is a scratch which is only one or two pixels wide. In such cases, optimal results can be produced after only a single iteration, which would render our algorithm several times faster again.

In spite of this, we found that loopy belief-propagation tended to converge in very few iterations. While we believe it helped that the regions we are inpainting appear to be fairly `tree-like', there is very little theory to support this claim. On the other hand, loopy belief-propagation often converges far slower when dealing with large regions, meaning that we can inpaint a `scratch' much faster than a `coffee stain'.

We have also not considered the possibility that the corrupted pixels may contain some information about the original image. Many gradient-ascent approaches implicitly exploit this possibility by initializing their algorithms using the corrupted pixels. If the restored image is `close to' the corrupted image, this can result in faster convergence. Our approach is also able to deal with this possibility by augmenting the graphical model with an observation layer with the respective noise model for the damaged pixels.

\section{Conclusion}

In this paper, we have developed a model for inpainting images quickly using belief-propagation. While image inpainting has previously been performed using low-order models by belief-propagation, and high-order models by gradient-ascent, we have presented new methods which manage to exploit the benefits of both, while avoiding their shortcomings. We have shown these algorithms to give satisfactory visual results and to be faster than existing gradient-based techniques, even in spite of our high-level implementation. 


\end{document}